\documentclass{article}

\usepackage{PRIMEarxiv}

\usepackage[utf8]{inputenc} 
\usepackage[T1]{fontenc} 

\usepackage{enumitem}
\usepackage{hyperref}       
\usepackage{url}            
\usepackage{booktabs}       
\usepackage{amsfonts}       
\usepackage{nicefrac}       
\usepackage{subfig}   
\usepackage{xcolor}
\usepackage{multirow} 
\usepackage{microtype}      
\usepackage{lipsum}
\usepackage{amsmath} 
\usepackage{fancyhdr}  
\usepackage{array}
\usepackage{pifont}
\usepackage{graphicx}       
\graphicspath{{media/}}     
\newcolumntype{C}{>{\centering\arraybackslash}p{1.2cm}}
\newcolumntype{L}{>{\raggedright\arraybackslash}p{2.8cm}}
\newcommand{\cmark}{\textcolor{black}{\ding{51}}} 
\newcommand{\xmark}{\textcolor{black}{\ding{55}}} 
\title{Text-conditioned State Space Model For  \\ Domain-generalized
Change Detection Visual Question Answering
}

\author{
  Elman Ghazaei \\
  Faculty of Engineering and Natural Sciences (VPALab),\\
  Sabanci University, T\"{u}rkiye\\
  \texttt{elman.ghazaei@sabanciuniv.edu} \\
   \And
  Erchan Aptoula \\
  Faculty of Engineering and Natural Sciences (VPALab), \\
  Sabanci University, T\"{u}rkiye\\
  \texttt{erchan.aptoula@sabanciuniv.edu} \\
}

\begin{document}
\maketitle

\begin{abstract}
Typical change detection methods produce change masks from bi-temporal images that require expert knowledge for accurate interpretation. The task of Change Detection Visual Question Answering (CDVQA) was introduced to alleviate this requirement.
However, so far reported CDVQA studies have been conducted under the assumption that training and testing datasets share similar distributions, which does not hold in real-world applications, where domain shifts often occur. That is why in this article the CDVQA task is studied from a domain generalization perspective with two major contributions. 
First, BrightVQA, a new multi-modal and multi-domain CDVQA dataset specifically constructed for domain generalization research, is introduced. Second, the Text-Conditioned State Space Model (TCSSM) is presented that has been designed to exploit jointly both bi-temporal imagery and publicly available geo-disaster-related textual information. Extensive experiments and ablation studies show that TCSSM leads to state-of-the-art results, systematically outperforming all counterparts.
The code and dataset will be made publicly available upon acceptance at github.com/Elman295/TCSSM.
\end{abstract}


\section{Introduction}
Change Detection (CD) is defined as the process of identifying regions of change over time, from multi-temporal images captured over the same geographic area \cite{lu2004change}, and constitutes a popular remote sensing image analysis task \cite{chen2024changemamba, zhang2025cdmamba}. CD enjoys a wide range of applications such as urban development planning \cite{10960557}, natural disaster management \cite{tian2014improving}, environmental monitoring \cite{10744421}, and agricultural analysis \cite{10364836}.

CD methods are classified into two broad classes: \textit{binary}, where regions are classified as ``changed'' or ``unchanged'', and \textit{semantic}, where changed areas are identified along with their specific type of change, thus providing more informative output \cite{chen2024changemamba}. However, the interpretation of CD masks at the pixel-level typically requires expert knowledge, making it difficult for end-users \cite{yuan2022change}. 
To address this limitation, visual question answering (VQA) \cite{antol2015vqa}, a task located at the intersection of natural language processing and computer vision, has been successfully adapted first to scene captioning in the form of remote sensing VQA (RSVQA) \cite{lobry2020rsvqa, yuan2022change, zhang2023spatial, zheng2021mutual}, and then recently to CD.
CDVQA thus enables users to naturally query for specific changes in multi-temporal images.

\begin{table*}[t]
\begin{center}
\small
\caption{Positioning of DG w.r.t.~its counterparts; 
$P^{S/T}_{XY}$: source/target joint distribution,
$\mathcal{Y}_{S/T}$: source/target label space, 
$P^T_{X/Y}$: target marginal. $^{\dagger}$ Limited in quantity, like a single example or mini-batch \cite{zhou2022domain}.}
\label{tab:cdvqa_settings}
\begin{tabular}{l|cc|cc|cc|c|c}
\toprule
 & \multicolumn{2}{c|}{\# of Domains}
 & \multicolumn{2}{c|}{$P^{S}_{XY}$ vs $P^{T}_{XY}$} 
 & \multicolumn{2}{c|}{$\mathcal{Y}_S$ vs $\mathcal{Y}_T$} 
 & Access to $P^T_X$ & Access to $P^T_Y$ \\

 & $=1$ & $>1$ & $=$ & $\neq$ & $=$ & $\neq$ & \\
\midrule
Supervised Learning   & \checkmark &        & \checkmark &        & \checkmark &        &      &  \\
Fine-tuning     & \checkmark & \checkmark &            & \checkmark &  & \checkmark & \checkmark &\checkmark\\
Zero-Shot Learning    & \checkmark &  &            & \checkmark &            & \checkmark &       & \\
Domain Adaptation     & \checkmark & \checkmark &            & \checkmark & \checkmark & \checkmark & \checkmark & \checkmark \\

Unsupervised Domain Adaptation     & \checkmark & \checkmark &            & \checkmark & \checkmark & \checkmark & \checkmark \\
Test-Time Adaptation    & \checkmark & \checkmark &            & \checkmark & \checkmark &  & \checkmark$^{\dagger}$ \\
\textbf{Domain Generalization} & \checkmark & \checkmark &            & \checkmark & \checkmark & \checkmark &        \\
\bottomrule
\end{tabular}
\end{center}
\end{table*}

Nevertheless, so far most, if not all, of RSVQA studies have been conducted under the assumption that training and testing data are drawn from similar distributions \cite{lobry2020rsvqa}. This widespread assumption is evidently unrealistic, since distribution shifts are commonly encountered \cite{zhou2022domain} (known as \textit{domain shift}), and can lead to significant performance degradations during deployment. This issue is especially aggravated in the case of CD and disaster-management, since not only two disasters are ever alike, but CDVQA can involve a wider variety of shifts (e.g.~question and answer shift) \cite{huang2025frames} than other RS tasks (e.g.~geographical shift, sensor shift, etc.).

Among the multiple strategies that have been proposed to address domain shift, Domain Adaptation (DA) along with its variants (unsupervised, source-free, test-time, etc) has been investigated intensively \cite{peng2022domain}. 
Yet, DA relies on the availability of target domain data, which is often an impractical (and sometimes outright impossible) requirement in real-world applications \cite{zhou2022domain} (e.g.~a company cannot know where its future customers are going to use its solution, or where the next disaster is going to strike, etc.). As a more realistic alternative, domain generalization (DG) assumes that neither data nor labels from the target domain are available during training \cite{unni2023vqa, zhang2021domain, liang2024single, zhang2023language} (Table \ref{tab:cdvqa_settings}). Under this setting, models are expected to generalize to unseen domains solely based on the knowledge learned from source/training domain(s). 

The focus of this article is the CDVQA task under a DG setting. To this end, a new model called Text-Conditioned State Space Model (TCSSM) based on state-of-the-art state space models (SSMs) is proposed. What sets apart TCSSM from existing studies is that not only it is capable of generating input-dependent parameters through the fusion of pre- \& post-event imagery, but it is designed to also exploit publicly available geo-disaster-related textual descriptions. This conditioning mechanism is hypothesized to allow the model to focus more effectively on domain-invariant features, via leveraging generic textual context (about the geography of countries, disasters, etc.) to adaptively guide parameter estimation. The motivation for this design stems from recent findings \cite{bose2024stylip} indicating that domain-invariant features in vision-language models (VLMs) can be better extracted by employing generic descriptive texts rather than instance-specific captions. 

In addition, and equally importantly, a new multi-modal dataset named BrightVQA has been constructed from the Bright dataset \cite{chen2025bright} specifically for DG oriented CDVQA. It contains 8 distinct question types, approximately 2.1 million question–answer pairs, 62 unique answers, and 54,224 bi-temporal image pairs spanning 9 countries (and 10 cities) rendering it a large-scale and diverse benchmark for evaluating CDVQA models under DG settings. 

The contributions of this article can be summarized as follows: 

1) This is the first study to define and explore DG in the context of CDVQA. It is a highly challenging task due to the variability of RS data across geographic regions, sensors, and change types such as floods, earthquakes, and urban developments. The only other reported CDVQA model \cite{yuan2022change} is prone to domain shift, limiting its applicability to unseen environments, while this study strives specifically for enhanced generalization capacity, so as to enable deployment in real-world scenarios.

2) Publicly available geo-disaster-related textual cues are incorporated into the proposed model to enhance generalization, enabling geography- and disaster-aware representation learning. These texts, collected from Wikipedia, are hypothesized to facilitate a deeper alignment between visual and textual modalities during feature extraction.

3) A novel state space model based architecture (TCSSM) is proposed, in which bi-temporal imagery and geo-disaster-related textual features are jointly utilized to predict input-dependent parameters. In contrast to traditional SSMs, where parameters are derived solely from visual or textual inputs, TCSSM enables dynamic parameter generation through cross-modal conditioning. Comprehensive experiments and ablation studies show that it achieves state-of-the-art results.

4) An automatic VQA dataset generation pipeline is designed, resulting in the \textit{BrightVQA} dataset. In comparison w.r.t.~the only other existing CDVQA dataset \cite{yuan2022change}, BrightVQA encompasses a greater variety of question types and incorporates two modalities - RGB and SAR - for pre- and post-event images. Furthermore, the dataset size is larger, and samples are distributed globally across multiple countries.

The remainder of this article is organized as follows. A review of related work in Section \ref{sec:realted} is followed by an overview of the proposed BrightVQA dataset (Section \ref{sec:dataset}). Next, TCSSM is introduced in Section \ref{sec:method}, and experiments are presented in Section \ref{sec:exp}, before the paper concludes with Section \ref{sec:conc}.

\section{Related Work} \label{sec:realted}

\subsection{Context and motivation}
In disaster monitoring tasks, datasets of bi-temporal data are commonly available from geographically and environmentally diverse regions, as disasters are sporadic and unpredictable. Hence, models trained on a single domain/region frequently fail when applied directly to unseen domains because of substantial distribution shifts \cite{zhou2022domain}. Consequently, DG constitutes an essential requirement for ensuring the practical utility and reliability of CDVQA systems in deployment. Performance degradations due to differences in terrain and topography can impede timely decision-making in disaster response operations, thus emphasizing the necessity of models capable of generalizing effectively across heterogeneous contexts.

Although CDVQA and DG have been examined independently \cite{yuan2022change,iizuka2023frequency, liang2024single, zhu2023style}, no prior research has explicitly investigated DG within the context of CDVQA. This gap highlights the absence of robust and generalizable systems capable of answering change-related questions across diverse and previously unseen real-world scenarios.

\subsection{Close-ended RSVQA}

Close-ended RSVQA constitutes essentially a classification problem where a predefined set of possible answers is used. It was introduced by Lobry et al.~\cite{lobry2020rsvqa}, along with two datasets and a baseline model consisting of a vision backbone (Resnet-152), a question backbone (recurrent neural network-RNN), a fusion module (Hadamard product), and a prediction head. Next, the RSIVQA dataset was presented by Zheng et al.~\cite{zheng2021mutual} and a new attention based fusion model was proposed. 

Soon after, the CDVQA task was introduced by Yuan et al.~\cite{yuan2022change} along with a new task specific dataset. Their model employed a convolutional neural network (CNN) and a RNN as vision and question backbones respectively, together with a multi-temporal and multi-modal fusion module. In another line of work \cite{bazi2022bi}, transformers were explored, where the roles of the key and value in the decoder were swapped, showing that such a modification can enhance VQA performance.

Furthermore, a question augmentation strategy was proposed in \cite{yuan2023multilingual} by translating repeatedly questions across multiple languages, thereby enriching the diversity of training samples. In a separate study \cite{yuan2022easy}, a self-paced learning strategy was presented for VQA, where a model was trained progressively from easy to hard examples. The use of object-level image features on the other hand was explored in \cite{felix2021cross}. Other prominent approaches include Prompt-RSVQA \cite{chappuis2022prompt}  that incorporated multi-label supervision and utilized DistilBERT~\cite{sanh2019distilbert} to better answer questions from the extracted labels. Next, RSAdapter \cite{wang2024rsadapter} was presented as a lightweight adaptation module that allows models to be fine-tuned without training from scratch. Finally, a new dataset along with a semantic label-aware model were introduced in \cite{wang2024earthvqanet}, in order to focus on relevant visual regions during reasoning. 

\subsection{Open-ended RSVQA}

In the open-ended setting, RSVQA is tackled as a generative problem, wherein answers are produced in free-form natural language without restriction to a predefined answer set. To address the scarcity of high-quality image–text datasets in remote sensing, Yuan et al.~\cite{hu2025rsgpt} introduced a human-annotated dataset specifically designed to facilitate the development of large VLMs. Building upon this, Liu et al.~\cite{liu2025rescueadi} proposed the novel task of Adaptive Disaster Interpretation, which is aimed at managing complex and multi-faceted disaster-related requests through the sequential execution of interrelated interpretation tasks, thereby enabling a comprehensive understanding of disaster scenarios. In a related development, Kuckreja et al.~\cite{kuckreja2024geochat} presented the first multitask conversational VLM tailored for remote sensing, capable of engaging in both image-level and region-specific dialogues. 
Moreover, Zhao et al.~\cite{zhao2024see} proposed a unified framework that simultaneously addresses semantic segmentation, image captioning, and VQA on high-resolution remote sensing images. More recently, Zhang et al.~\cite{zhang2024earthgpt} employed a unified instruction tuning method to handle diverse remote sensing tasks, including scene classification, image captioning, VQA, and visual grounding. 

Despite these advances, all aforementioned closed-ended and open-ended RSVQA methods have been developed under the assumption that the training and testing data are drawn from similar distributions. As such, the challenge of domain shift remains unaddressed in prior RSVQA research.

\subsection{Overview of RSVQA datasets}
Table \ref{tab:datacomp} summarizes the characteristics of existing datasets in comparison to the proposed dataset.

\begin{table*}[ht]
\begin{center}
\scriptsize
\caption{Comparison of the proposed BrightVQA dataset against benchmark RSVQA datasets. The symbol ``--'' highlights the absence of the corresponding information in the original paper.}
\label{tab:datacomp}
\begin{tabular}{lcccccccc}
\toprule
\textbf{Dataset}  & \textbf{Images} & \textbf{Questions} & \textbf{Question} & \textbf{Image} & \textbf{Countries} & \textbf{Modality} & \textbf{Supports} & \textbf{Supports} \\
& & & \textbf{Types} & \textbf{Size} & & & \textbf{CD} & \textbf{DG} \\
\midrule
LR RSVQA (TGRS'20) \cite{lobry2020rsvqa} & 772 & 77,232 & 4& 256$\times$256 & Netherlands & RGB& \ding{55}&\ding{55} \\

HR RSVQA (TGRS'20) \cite{lobry2020rsvqa} & 10,659 & 1,066,316 &4 & 512$\times$512 & USA  &RGB &\ding{55}&\ding{55}\\

RSVQAxBen (IGARSS'21) \cite{lobry2021rsvqa} &  590,326& 14,075,801 & 2 & 120$\times$120 &  -- & RGB &\ding{55}&\ding{55}\\

RSIVQA (TGRS'22) \cite{zheng2021mutual} & 37,264 & 111,134 &3 & -- & -- & RGB&\ding{55}& \ding{55}\\

CDVQA (TGRS'22) \cite{yuan2022change} & 2,968  & 122,000 & 4 & 512$\times$512 & China & RGB &\ding{51}&\ding{55}\\

TextRS-VQA (JSTAR'23) \cite{bashmal2023visual} & 2,144 & 6,245 & 4 & 256$\times$256 & -- & RGB&\ding{51}&\ding{55}\\

EarthVQANet (ISPRS'24) \cite{wang2024earthvqanet} & 6,000   & 208,593 & 6 & 1,024$\times$1,024 & China & RGB &\ding{55}&\ding{51}\\

\midrule

\textbf{BrightVQA}  & {54,224} & 2,168,960 & {8} & 256$\times$256 & 
Congo, Equatorial Guinea,  & RGB, SAR & \ding{51}&\ding{51} \\
& & & & & 
Haiti, Lebanon, Libya & & \\
& & & & & 
Morocco, Spain, Türkiye, USA & & \\
\bottomrule
\end{tabular}
\end{center}
\end{table*}

\textbf{Low Resolution RSVQA} \cite{lobry2020rsvqa}: is a Sentinel-2 based dataset containing images from the Netherlands, with 10 m resolution. A total of 9 low cloud cover tiles were divided into 772 image patches of size 256 $\times$ 256 pixels. From these, 77,232 question–answer pairs were generated, spanning 4 question types: count, rural/urban classification, comparative, and presence.

\textbf{High Resolution RSVQA} \cite{lobry2020rsvqa}: is a RGB dataset using 15 cm resolution aerial imagery from the USGS High-Resolution Orthoimagery collection, covering urban areas across the USA. 161 tiles from the North-East coast were divided into 10,659 images of size 512 $\times$ 512 pixels. A total of 1,066,316 question–answer pairs were generated, spanning 4 question types: count, comparative, area, and presence.

\textbf{RSVQAxBen} \cite{lobry2021rsvqa}: is another Sentinel-2 based dataset containing 590,326 RGB image patches at 10 m resolution. A set of 25 diverse questions grounded in land cover semantics and spatial characteristics were generated per image, resulting in a total of 14,075,801 image–question–answer triplets.

\textbf{RSIVQA} \cite{zheng2021mutual}: has been constructed by combining five aerial scene classification datasets, with both automatically generated and human-annotated 111,134 VQA triplets, categorized into 3 answer types: yes/no, numerical, and other. 

\textbf{CDVQA} \cite{yuan2022change}: Yuan et al.~constructed this dataset based on SECOND (SEmantic Change detectiON Dataset) \cite{yang2020semantic}. It consists of bi-temporal RGB aerial images acquired from various sensors, with spatial resolutions ranging from 0.5 to 3 m. It covers several Chinese cities such as Shanghai, Hangzhou, and Chengdu. It contains 4,662 image pairs (512 $\times$ 512 pixels), of which 2,968 pairs are publicly available. Each pair includes pre- and post-event images, along with their corresponding semantic change maps annotated at pixel level across 6 classes: non-vegetated ground, buildings, playgrounds, water, low vegetation, and trees.

\textbf{TextRS-VQA} \cite{bashmal2023visual}: was constructed based on images from the TextRS dataset \cite{abdullah2020textrs}, originally introduced for image retrieval and captioning tasks. It comprises 2,144 images with various spatial resolutions and image dimensions and is manually annotated with 2 to 5 questions, resulting in a total of 6,245 questions spanning 4 categories: object counting, presence/absence, class type, and a diverse ``other'' category. 

\textbf{EarthVQANet} \cite{wang2024earthvqanet}: was developed as an extension of the LoveDA dataset \cite{wang2021loveda}, which covers 18 urban and rural regions across Nanjing, Changzhou, and Wuhan. It provides 6,000 0.3 m resolution images along with semantic masks for 7 land-cover classes. It contains 208,593 question–answer pairs focusing mostly on city planning. Each urban and rural image is paired with 42 and 29 question-answer pairs respectively.

\section{The BrightVQA dataset} \label{sec:dataset}


\begin{table*}[t]
\small
\begin{center}
\caption{Examples of question-answer pairs for each question type in the BrightVQA dataset.}
\label{tab:brightvqa_samples}
\begin{tabular}{c|c|c|c}
\toprule
Question Category & Question & Answer & Number of \\ 
& & & Questions\\ 
\midrule
Damage Detection & Is there any \textbf{visible damage} in the post-disaster image? & Yes/No & 325,344\\ 
 Quantitative& What \textbf{percentage} of the area shows damage or destruction? & 0-10\%  & 433,792\\ 
Comparative & Which \textbf{type of damage} is more prevalent: \textbf{damaged or destroyed buildings}? & Damaged  & 325,344\\ 
Severity & How would you classify the \textbf{overall damage severity}? & Minor Damage  & 216,896\\ 
Spatial & Is the damage \textbf{concentrated in one area or spread throughout}? & Concentrated  & 108,448\\ 
Contextual & How \textbf{effective were the buildings} in Türkiye at withstanding the earthquake? & Very Effective  & 216,896\\ 
Threshold & Is \textbf{less than 5\%} of the area damaged? & Yes/No  & 325,344\\ 
Recovery Assessment & Will this area \textbf{require extensive reconstruction}? & Yes/No  & 216,896\\ 
\bottomrule
All & & & 2,168,960\\ 
\end{tabular}
\end{center}
\end{table*}

The recently introduced semantic change detection dataset Bright \cite{chen2025bright} has been chosen as the basis for automatically generating the proposed new CDVQA dataset named \textit{BrightVQA}. It contains optical RGB pre-event and SAR post-event images with spatial resolutions ranging from 0.3 to 1 m. This fixed temporal-modal combination is designed to reflect a realistic disaster-response scenario, where information-rich optical imagery is regularly acquired pre-event, and SAR imagery, being less affected by cloud cover, is more readily accessible immediately post-event. The bi-temporal images and their corresponding masks originate from multiple countries thus rendering the dataset suitable for the DG setting (Table \ref{tab:datacomp}). 
It possesses 3,389 labeled pairs of multi-modal images, each with a size of 1,024 $\times$ 1,024 pixels, that were cropped into tiles of size 256 $\times$ 256 pixels, leading to a total of 54,224 pairs. A single semantic mask accompanies each of them, with 4 human-verified labels: intact, destroyed, damaged, and background. Moreover, BrightVQA contains exactly 40 question–answer pairs per sample. The questions are fixed for each image pair, and the answers are automatically constructed based on the available image–mask information. These questions are categorized into 8 semantic types: damage detection, quantitative, comparative, severity, spatial, contextual, threshold, and recovery assessment that are presented next (Table~\ref{tab:brightvqa_samples}).

\subsection{Multitemporal Question–Answer pair Construction}
Let $f^{pre} \in \mathbb{R}^{C_{pre} \times H \times W}$ and $f^{post} \in \mathbb{R}^{C_{post} \times H \times W}$ be the pre-event and post-event images, where $C_{pre/post}$, $H$ and $W$ represent their number of channels, height and width respectively. Each image pair is associated with a human-verified single mask $s \in \mathbb{N}^{H \times W}$, and its every pixel can take one of the following values: 0 (``background''), 1 (``intact''), 2 (``damaged''), or 3 (``destroyed''), describing the semantic state change during the observed event. Details about each question type follow.

\textbf{Damage detection}: aims to identify the presence of damage, destruction, or intact structures. The number of pixels corresponding to a specific label $l \in \{0,1,2,3\}$ is computed as  
$
Q_l = \sum_{i=1}^H \sum_{j=1}^W \mathbf{1}(s_{i,j} = l),$ where $\mathbf{1}(\cdot)$ is the indicator function. For each label $l$, a binary (Yes/No) answer is generated based on whether $Q_l > 0$ or not. For example, the presence of damaged structures is indicated by:  
\begin{equation}
\text{Answer} = 
\begin{cases}
\text{Yes}, & \text{if } Q_2 > 0, \\
\text{No}, & \text{otherwise}.
\end{cases}
\end{equation}

\textbf{Quantitative}: quantifies the extent of damage, destruction, or intact areas of each class $l$ as a pixel percentage w.r.t.~either the entire scene or the total number of building pixels:
\begin{equation}
\left\lfloor 100 \times \frac{Q_l}{H \times W} \right\rfloor \text{ or } \left\lfloor 100 \times \frac{Q_l}{Q_1 + Q_2 + Q_3} \right\rfloor
\end{equation}
The percentages are then converted into categorical intervals (e.g.~``0–10\%'', ``10–20\%'') for answer generation. 

\textbf{Comparative}: focuses on the type of dominant destruction in a given sample, by comparing their pixel coverage; e.g. the dominant destruction type between ``damaged'' and ``destroyed'' is determined by comparing $Q_2$ vs $Q_3$:  
\begin{equation}
\text{Answer} = 
\begin{cases}
\text{Damaged}, & \text{if } Q_2 > Q_3, \\
\text{Destroyed}, & \text{otherwise}.
\end{cases}
\end{equation}

\textbf{Severity}: Assesses the overall severity $(os)$ of destruction ($l=2$ or $l=3$) in the scene based on the proportion of affected areas:
\begin{equation}
os = \left\lfloor 100 \times \frac{Q_2 + Q_3}{H \times W}\right\rfloor
\end{equation}

It is then subdivided into multiple severity levels $(sl_i), i \in \{1,2,3,4,5\}$ based on predefined thresholds:
\begin{equation}
\begin{split}
sl_1 = & \text{ No damage}, \text{if } os = 0, \\
sl_2 = & \text{ Minor damage}, \text{if } 0 < os < 10, \\
sl_3 = & \text{ Moderate damage}, \text{if } 10 \leq os < 30, \\
sl_4 = & \text{ Severe damage}, \text{if } 30 \leq os < 60, \\
sl_5 = & \text{ Extensive damage}, \text{otherwise.}
\end{split}
\end{equation}

To validate these thresholds, the dataset's $sl$ distribution was analyzed (Table \ref{tab:freq_freq}), and according to \cite{zhu2020adjusting}, it exhibits a reasonable level of balance, with an imbalance ratio of 2.55.

\begin{table}[h]
\caption{Severity Level Frequencies}
\label{tab:freq_freq}
\begin{center}
\begin{tabular}{c|ccccc}
\toprule
Severity Level       & $sl_1$   & $sl_2$   & $sl_3$ &$sl_4$ &$sl_5$   \\
\midrule
Frequency & 0.18 & 0.22 & 0.28 & 0.21 & 0.11 \\
\bottomrule
\end{tabular}
\end{center}
\end{table}

\textbf{Spatial}: evaluates the spatial distribution of destruction within an image, distinguishing between concentrated and widespread destruction patterns. A binary destruction mask is created by selecting pixels labeled as damaged or destroyed ($l = 2$ or $l = 3$). The Euclidean distances of these pixels from their centroid are computed and compared to their standard deviation $\sigma$.
If more than 70\% of the distances are below this threshold, then the distribution is labeled as ``concentrated in one area''; otherwise, it is considered ``spread throughout''.

The chosen threshold yields a balanced distribution with 52\% of samples classified as ``concentrated'' and 48\% as ``spread throughout'' ensuring adequate representation of both destruction patterns in the BrightVQA dataset. 

\textbf{Contextual}: provides disaster-specific insights, focusing on structural resilience ($sr$):
\begin{equation}
sr = \left\lfloor 100 \times \frac{Q_{1}}{Q_{1} + Q_{2} + Q_{3}} \right\rfloor
\end{equation}
It is then subdivided into multiple resilience levels $(r_i), i \in \{1,2,3,4,5\}$ based on predefined thresholds:
\begin{equation}
\begin{split}
r_1 = & \text{ Very effective}, \text{if } sr \geq 80, \\
r_2 = & \text{ Moderately effective}, \text{if } 60 \leq sr < 80, \\
r_3 = & \text{ Mixed effectiveness}, \text{if }  40\leq sr < 60, \\
r_4 = & \text{ Poor effectiveness}, \text{if } 20 \leq sr < 40, \\
r_5 = & \text{ Very poor effectiveness}, \text{otherwise.}
\end{split}
\end{equation}
To validate these thresholds, the dataset's $r$ distribution was analyzed (Table \ref{tab:freq_freq_2}) and it exhibits an imbalance ratio of 2.7 which is acceptable according to \cite{zhu2020adjusting}.

\begin{table}[h]
\caption{Resilience Level Frequencies}
\label{tab:freq_freq_2}
\begin{center}
\begin{tabular}{c|ccccc}
\toprule
Resilience Level       & $r_1$   & $r_2$   & $r_3$ &$r_4$ &$r_5$   \\
\midrule
Frequency & 0.27 &0.21 & 0.23 & 0.19 & 0.10  \\
\bottomrule
\end{tabular}
\end{center}
\end{table}

\textbf{Threshold}: assesses the extent of damage in an area via binary (yes/no) questions based on whether the destruction ratio ($p_{dest}$):
\begin{equation}
p_{dest} = \left\lfloor 100 \times \frac{Q_{2} + Q_{3}}{H \times W} \right\rfloor
\end{equation}
\noindent exceeds or falls below a list of predefined critical levels: 5\%, 10\%, 25\%, 50\%, 75\%, and 90\%.


\textbf{Recovery Assessment}: evaluates the binary necessity for reconstruction or emergency services based on the extent of destruction. Reconstruction need is indicated if $p_{dest} > 40\%$. Four questions evaluate reconstruction needs, emergency service requirements, and habitability. The threshold of 40\% was chosen for the sake of distribution balance, leading to a 49\%-51\% distribution of possible answers.

\begin{figure*}[h]
    \begin{center}
    \includegraphics[width=\linewidth]{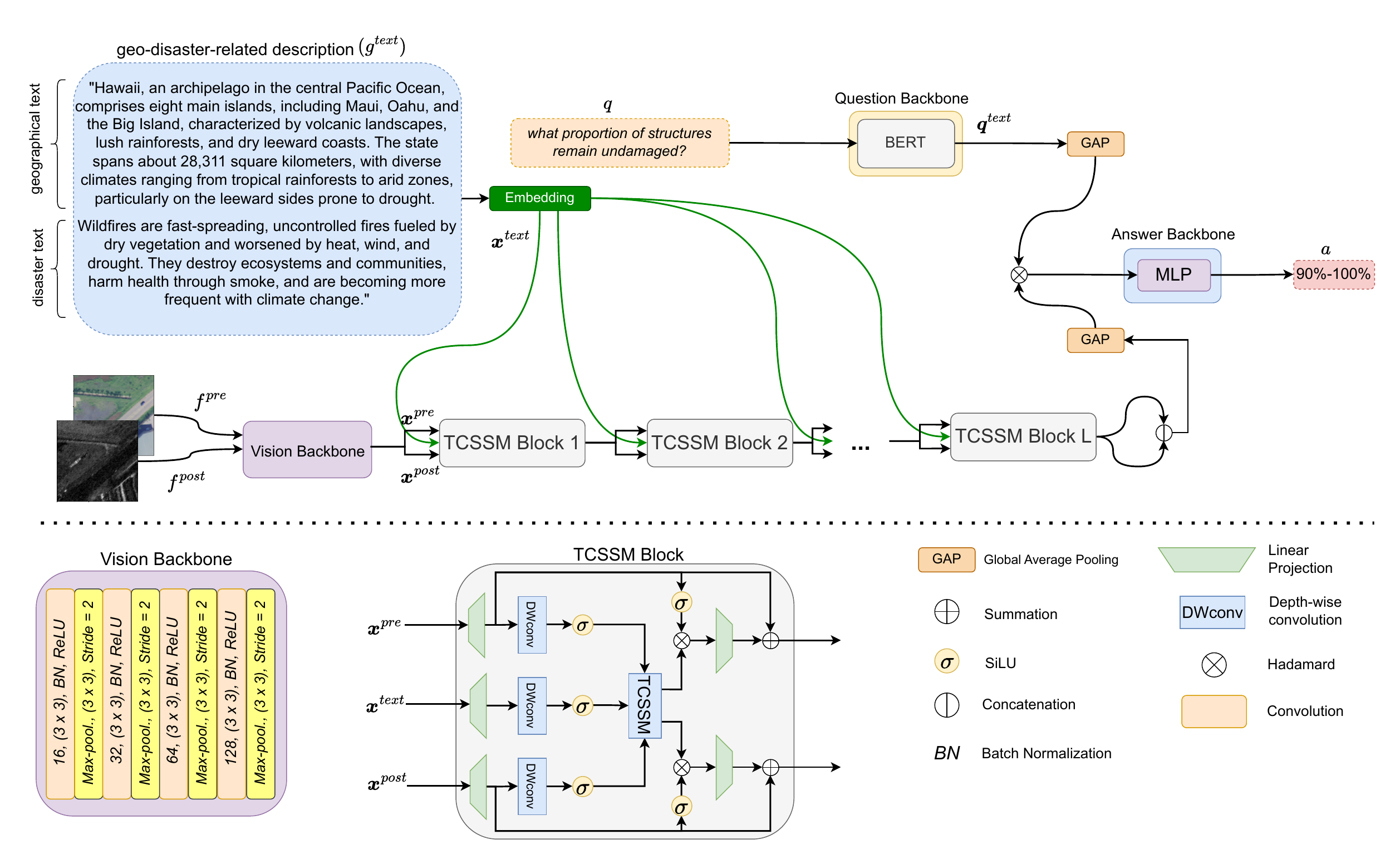}
    \caption{Illustration of the proposed TCSSM model.}
    \label{fig:tcssm_main}
    \end{center}
\end{figure*}

\subsection{Question Distributions}
The distribution of the 8 question categories in the BrightVQA dataset is presented in Table \ref{tab:brightvqa_samples}. Each of them is designed to probe a different aspect of reasoning within the context of disaster-related visual information. Among them, quantitative questions dominate the distribution, accounting for 433,792 instances. These questions typically involve numerical reasoning, such as estimating damage percentages or comparing the extent of destruction. They are followed by damage detection, comparative, and threshold categories, each contributing 325,344 questions. These categories often require binary or ordinal reasoning, such as detecting whether damage occurred, comparing severity levels, or determining if damage exceeds a certain threshold.

\section{A Text-Conditioned State Space Model}
\label{sec:method}


\subsection{Problem Definition and TCSSM Overview}
CDVQA is essentially a classification task \cite{lobry2020rsvqa, zheng2021mutual, wang2024earthvqanet, wang2024earthvqa, wang2024rsadapter} and its objective consists in providing natural language answers to questions about changes observed between a pair of bi-temporal images. Each data sample is composed of a pre-event image, a post-event image, and a corresponding question--answer pair. The proposed TCSSM has been designed for this task, with a particular emphasis on generalization capacity.

Let $\mathcal{D_S} = \{ \mathcal{D}_S^{(j)}\}_{j=1}^{N_S}$ and $\mathcal{D_T} = \{ \mathcal{D}_T^{(j)}\}_{j=1}^{N_T}$ denote the sets of source and target domains respectively. All domains are assumed to (potentially) possess distinct marginal distributions. In the DG setting, only $\mathcal{D_S}$ is accessible during training. Each of the data domains is represented as $\mathcal{D}_{S/T}^{(j)} = \{(f^{pre}_{i}, f^{post}_{i}, q_i, a_i)\}_{i=1}^{N_{S/T}^{(j)}}$, where $f^{pre}_{i}$, $f^{post}_{i}$, $q_i$, and $a_i$ denote the pre-event image, post-event image, natural language question and its answer, respectively. 
The goal of DG in CDVQA is to ensure that a model $m: (f^{pre}, f^{post}, q, \boldsymbol{\theta}) \rightarrow a$, with parameters $\boldsymbol{\theta}$ maximally exploits the various source domains in $\mathcal{D}_S$ and their inter-relationships so as to generalize effectively to $\mathcal{D}_T$ during testing, with no access to target domain(s) during training.

This study extends the model definition by including one more input $(g^{text})$ in $m$, constituting a publicly accessible joint textual description of the geography and of the disaster that has struck the depicted region.

As illustrated in the overview (Fig.~\ref{fig:tcssm_main}), the proposed TCSSM model receives a question, a geo-disaster-related description, and bi-temporal images as inputs, and predicts an answer as output. Specifically, the architecture comprises four main components: i) a vision backbone extracting features from the bi-temporal images; ii) a question backbone encoding the semantic representation of the input question; iii) the proposed TCSSM blocks, sequentially connected, designed to extract domain-invariant features and iv) a head producing the final output. Details of each component follow.

\subsection{Vision Backbone}
The vision backbone's task is to featurize the heterogeneous bi-temporal images $f^{pre}$ and $f^{post}$. Due to the significant modality gap between color and SAR images (commonly employed for disaster management oriented CD), each of them is processed independently using the same convnet (SAR-based processing is not the main focus of this study, as a standard convnet is employed for feature extraction). This parallel design allows the network to learn modality-specific features while preserving architectural consistency for downstream fusion.
The number of channels is progressively increased through the layers 
enabling the extraction of fine-grained and abstract representations. This separate yet symmetric processing scheme is designed to prepare both modalities for effective cross-temporal comparison and subsequent integration within the model.

\subsection{Question Backbone}
The question backbone's task is to extract contextualized representations from natural language questions. BERT \cite{devlin2019bert} has been adopted for this role, due to its strong semantic and syntactic information capturing performance. Specifically, a pre-trained BERT-base model has been used to encode each question into a sequence of contextualized token embeddings. To further tailor BERT for the CD task, a fully connected projection layer followed by a ReLU activation function has been appended, thus enabling the transformation of BERT’s high-dimensional output into a feature space compatible with the other modalities.


\begin{figure*}[t]
  \begin{center}
  \subfloat[Selection Mechanism in Mamba \cite{gu2023mamba}]
  {\label{fig.example1}\includegraphics[width=0.37\textwidth]{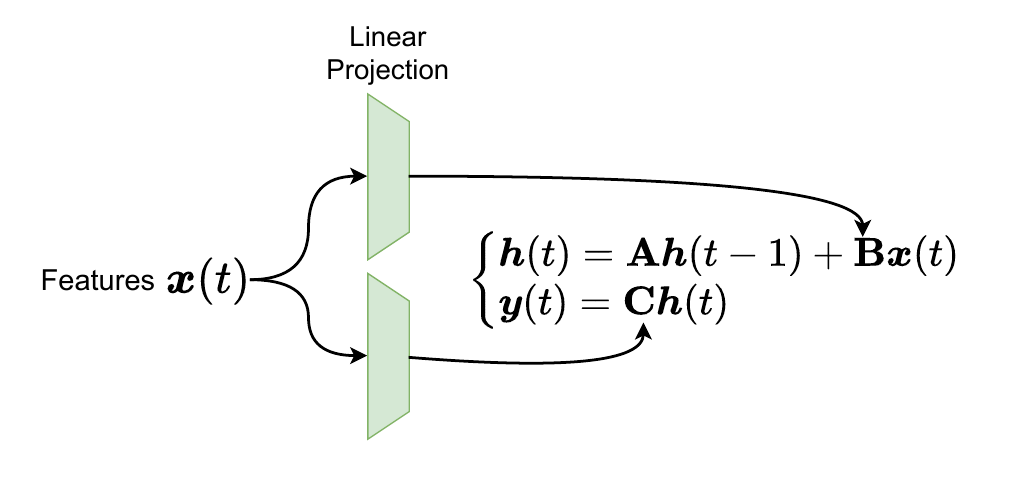}}
  \subfloat[Selection Mechanism in the proposed TCSSM]{\label{fig.example2}\includegraphics[width=0.60\textwidth]{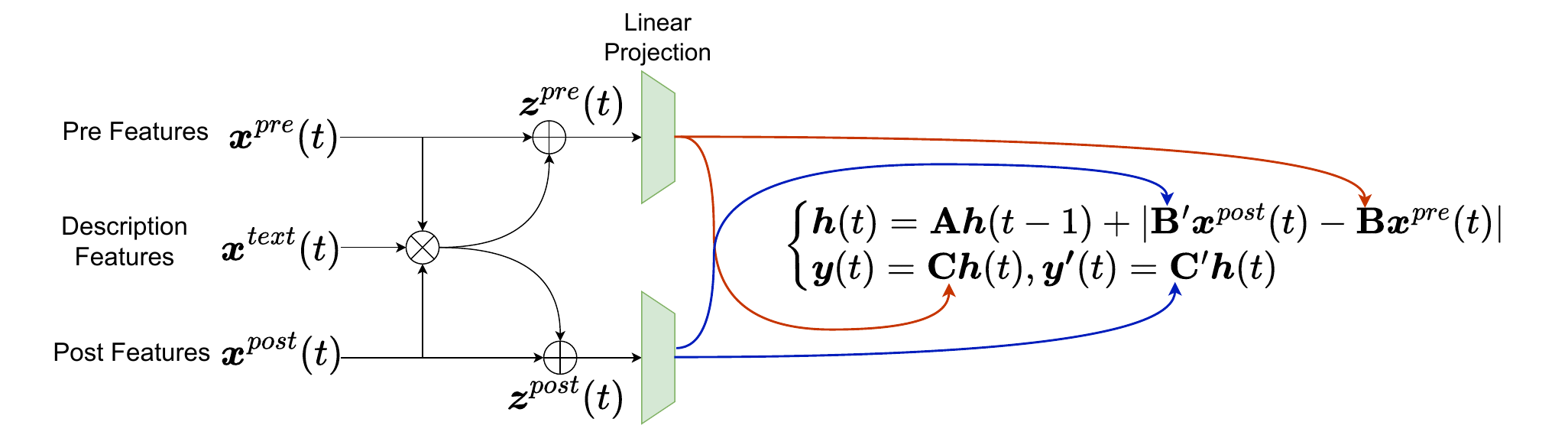}}
  \caption{Comparison of selection mechanisms in Mamba vs the proposed TCSSM.}
  \label{fig:abbb}
\end{center}
\end{figure*}

\subsection{TCSSM block}


\textbf{Mamba:} 
is a selective type of SSM, initially proposed \cite{gu2023mamba} as an efficient alternative to transformers that suffer from quadratic complexity w.r.t.~their input. Mamba is capable of long-range dependency modeling and input-dependent parameter estimation, while still enjoying linear complexity.

It is inspired by linear time-invariant systems, which typically map a sequence $\boldsymbol{x}(t) \in \mathbb{R}^d$ to an output $\boldsymbol{y}(t) \in \mathbb{R}^M$ through an implicit latent state $\boldsymbol{h}(t) \in \mathbb{R}^N$. These systems are usually formulated as linear ordinary differential equations: 
\begin{equation}
\begin{aligned}
\boldsymbol{h}(t) & = {\mathbf{A}} \boldsymbol{h}(t-1) + \mathbf{B} \boldsymbol{x}(t) \\
\boldsymbol{y}(t) & = \mathbf{C} \mathbf{h}(t)
\end{aligned} \xrightarrow[\text{ZOH}]{\text{discretize}}
\begin{aligned}
\mathbf{h}_t & = \overline{{\mathbf{A}}} \mathbf{h}_{t-1} + \overline{{\mathbf{B}}} x_t \\
y_t & = \mathbf{C} \mathbf{h}_t
\end{aligned}
\label{eq:mamba_rec}
\end{equation}
\noindent where for a hidden state vector $\boldsymbol{h}$, $\mathbf{{A}} \in \mathbb{R}^{N \times N}$ denotes the state transition matrix (i.e.~how the state evolves on its own), and $\mathbf{B} \in \mathbb{R}^{N \times d}$ (i.e.~how the input affects the state), $\mathbf{C} \in \mathbb{R}^{M}$ (i.e.~how the state becomes output), are respectively the input and output projection matrices.  In order to discretize SSMs, a time scale parameter $\Delta$ (typically a scalar) is employed commonly via a zero-order hold (ZOH) technique:
\begin{equation}
\overline{\mathbf{A}} = \exp(\Delta \mathbf{A}), \quad
\overline{\mathbf{B}} = (\Delta \mathbf{A})^{-1} \big( \exp(\Delta \mathbf{A}) - I \big)\, \Delta \mathbf{B}
\label{eq:zoh_ab}
\end{equation}

$\mathbf{A}$, $\mathbf{B}$, $\mathbf{C}$ and $\Delta$ are fixed in SSMs and are thus limited in terms of their capacity to adapt to their input. Mamba's main novelty is employing input-dependent ${\mathbf{B}_t}, \mathbf{C}_t, \Delta_t$ in Eqs.~\eqref{eq:mamba_rec} and \eqref{eq:zoh_ab} that are generated dynamically from the input sequence via learnable linear projections (Fig.~\ref{fig.example1})
thus allowing the model to adapt its computations in an input-dependent manner and improve its ability to model complex temporal patterns.


\textbf{A Domain-Invariant Mamba:} It has been reported \cite{bose2024stylip} that leveraging generic descriptive text leads to domain invariant features in VLMs. For instance, in the case of CLIP, such generic descriptions in place of image-specific captions have been shown \cite{radford2021learning} to enhance generalization. In the context of VQA, however, the questions cannot be relied upon for extracting domain-invariant information, as they typically lack explicit and transferable semantic content about the visual input. This limitation is particularly evident in disaster scenarios. 

As an alternative, it is proposed to use publicly available geo-disaster-related textual descriptions, ensuring universal availability and reproducibility. First, the geographical description of the affected country/region is collected via its corresponding Wikipedia section \cite{aydin2025domain}; then, general textual descriptions of the disaster type are gathered again from Wikipedia\footnote{Prompt and samples from Wikipedia are provided in the GitHub repository.}. Subsequently, a LLM (Grok) was employed to not only coherently merge these two types of text, but also to identify and correct any inconsistencies or errors via prompts. The final output $(g^{text})$ was used as the generic descriptive text for conditioning the proposed model (Fig.~\ref{fig:tcssm_main}).

Contrary to Mamba where its parameters are typically learned only from visual input, in the proposed TCSSM they are dynamically generated by conditioning on both the bi-temporal images and the corresponding geo-disaster-related description. Thus the model is hypothesized to be rendered contextually geo- and disaster-aware. This cross-modal conditioning is designed to guide the network toward capturing domain-invariant representations, thereby presumably enhancing generalization in VQA tasks across varying locations.

\textbf{TCSSM selection mechanism}: 
Contrary to Mamba's selection mechanism, 
given the pre-event ($\boldsymbol{x}^{pre}$) and post-event ($\boldsymbol{x}^{post}$) features, obtained from the vision backbone, 
the L1 metric $|\cdot|$ is used to compute the distance between them:
\begin{equation}
\begin{aligned}
\mathbf{h}(t) & = \mathbf{A} \mathbf{h}(t-1) + |\mathbf{B'}_t \boldsymbol{x}^{post}(t) - \mathbf{B}_t \boldsymbol{x}^{pre}(t)| \\
\boldsymbol{y}(t) & = \mathbf{C}_t \mathbf{h}(t), \quad 
\boldsymbol{y}'(t) = \mathbf{C'}_t \mathbf{h}(t) 
\end{aligned}
\label{eq:cssm_formula_dis}
\end{equation}
\noindent where 
$\mathbf{B}' \in \mathbb{R}^{N \times d}$
and $\mathbf{C}' \in \mathbb{R}^{N}$ are two additional \textit{projection parameters}, computed once again via linear projections of the embeddings.

Moreover, instead of using linear projections of a single input modality $(i.e.~\boldsymbol{x}^{pre}(t),\boldsymbol{x}^{post}(t))$, TCSSM 
employs for the same task the fusion $(\boldsymbol{z}^{pre}(t),\boldsymbol{z}^{post}(t))$ of geo-disaster text features $\boldsymbol{x}^{text}$ as produced by the question backbone, with the pre- \& post-event visual features through a Hadamard product ($\times$) (Fig.~\ref{fig.example2}):
\begin{equation}
    \boldsymbol{z}^{pre}(t) =  \boldsymbol{x}^{pre}(t) \times \boldsymbol{x}^{post}(t) \times \boldsymbol{x}^{text}(t) + \boldsymbol{x}^{pre}(t)
    \label{xpre}
\end{equation}
\begin{equation}
    \boldsymbol{z}^{post}(t) =  \boldsymbol{x}^{pre}(t) \times \boldsymbol{x}^{post}(t) \times \boldsymbol{x}^{text}(t) + \boldsymbol{x}^{post}(t)
    \label{xpost}
\end{equation}
in order to generate the input-dependent parameters $\mathbf{B}_t$, $\mathbf{B'}_t$, $\mathbf{C}_t$ and $\mathbf{C'}_t$. It is hypothesized that this combined use of the two modalities enables TCSSM to learn a structured alignment between visual representations (pre- and post-event features) and semantic knowledge derived from geo-disaster-related text. By enforcing this multimodal alignment, TCSSM can presumably not only capture finer correlations between event imagery and textual context but also enhance its robustness to domain shifts. 
The Hadamard product has been chosen due to its predominant use in multi-modal fusion, and for its ability to capture non-linear relationships between different modalities while maintaining linear computational complexity \cite{chrysos2025hadamard}.


After the application of the ZOH technique as per Eq.~\eqref{eq:zoh_ab}  


\noindent we obtain the discretized TCSSM system:
\begin{equation}
\begin{aligned}
&\mathbf{h}_t = \overline{\mathbf{A}} \mathbf{h}_{t-1} + |\overline{\mathbf{B'}} \boldsymbol{x}^{post}_{t} - \overline{\mathbf{B}} \boldsymbol{x}^{pre}_{t}| \\
&\boldsymbol{y}_t = \mathbf{C} \mathbf{h}_t, \quad \boldsymbol{y'}_t = \mathbf{C'} \mathbf{h}_t 
\end{aligned}
\label{eq:tcssm_formula_dis}
\end{equation}
\noindent where $\overline{\mathbf{B}}$, $\mathbf{C}$ and $\overline{\mathbf{B'}}$, $\mathbf{C}'$ are linear projections of $\boldsymbol{z}^{pre}(t)$ and $\boldsymbol{z}^{post}(t)$ respectively (Fig.~\ref{fig.example2}).

\textbf{TCSSM as a block}: each of the sequentially connected $L$ TCSSM blocks (Fig.~\ref{fig.example2}) receives three inputs: pre-event and post-event image features, as well as geo-disaster-related textual embeddings, and processes them using a sequence of linear projections, depthwise convolutions (inspired by \cite{liu2024vmamba}), configured with identical input and output channel dimensions, SiLU activation functions and of course the TCSSM selection mechanism. Each TCSSM block produces two outputs ($\boldsymbol{y}_t$ and $\boldsymbol{y'}_t$), which replace respectively $\boldsymbol{x}^{pre}_t$ and $\boldsymbol{x}^{post}_t$ in the following block.

\subsection{Answer Prediction Backbone}
For answer prediction, question features ($\boldsymbol{q}^{text}$) and fused vision and text features (coming from TCSSM) are sampled via Global Average Pooling, and then fused by Hadamard product inspired by \cite{chrysos2025hadamard}. Finally, a MLP is employed to predict the correct answer from a predefined set of 62 possible answer classes present in the dataset.



\section{Experiments} 
\label{sec:exp}
\subsection{Datasets and Implementation Details}
The proposed TCSSM was validated primarily on the proposed BrightVQA dataset. In the context of BrightVQA a leave-one-domain-out strategy was employed across the ten available regions/domains. All experiments were conducted on a NVIDIA RTX 4090 GPU. The Adam optimizer~\cite{kingma2014adam} was employed with an initial learning rate of $1 \times 10^{-4}$, weight decay of $1 \times 10^{-5}$, and a batch size of 32. All models were trained via a categorical cross-entropy loss for 100 epochs, with the learning rate decayed by a factor of 0.1 every 5 epochs. Augmentations were applied only to visual inputs, consisting of horizontal and vertical flips and pixel intensity normalization. Performance evaluation was measured using through overall accuracy (OA), average accuracy (AA), and F1-score.

\begin{table*}[t]
\begin{center}
\scriptsize
\caption{CDVQA performances on the BrightVQA dataset; the best result per region is highlighted in \textcolor{red}{red}, and the second-best in \textcolor{blue}{blue}.}
\label{tab:transposed_performance}
\begin{tabular}{llccccccc|c}
\toprule
\textbf{Location}& \textbf{Metric} & \textbf{RSVQA}   & \textbf{RSIVQA}   & \textbf{CDVQA} & \textbf{BiModal} & \textbf{SOBA}  & \textbf{RSAdapter}   & \textbf{EarthVQANet}  & \textbf{TCSSM}\\

&  & (TGRS'20)  & (TGRS'22)  & (TGRS'22) & (TGRS'22) & (AAAI'24)  & (TGRS'24)  & (ISPRS'24) & (Ours) \\
\midrule
\multirow{2}{*}{\textbf{Bata}}
&{OA(\%) } & 82.63 & 82.76 & 84.53 & 84.96 & 83.12 & \textcolor{blue}{\textbf{86.47}} & 83.28 & \textcolor{red}{\textbf{86.87}} \\

 & AA (\%) &78.00&79.01&86.59&\textcolor{blue}{\textbf{86.81}} &81.30&\textcolor{red}{\textbf{87.95}}&83.38&85.94 \\

&{F1(\%) } & 26.51 & 26.74 & 25.66 & 24.47 & 27.37 & \textcolor{blue}{\textbf{28.70}} & 27.80  & \textcolor{red}{\textbf{29.66}} \\

\midrule
\multirow{2}{*}{\textbf{Beirut}}
&{OA(\%) } & 89.02 & 89.54 & 86.76 & 84.37 & 90.14 & 91.33 & \textcolor{blue}{\textbf{91.47}} & \textcolor{red}{\textbf{92.31}} \\

 & AA (\%)  &90.58&89.92&87.79&85.10&90.70&\textcolor{red}{\textbf{92.61}}&91.25&\textcolor{blue}{\textbf{92.38}}\\
 
&{F1(\%) } & 28.78 & 29.10 & 24.87 & 22.51 & 29.47 & \textcolor{blue}{\textbf{30.64}} & 30.19   & \textcolor{red}{\textbf{34.00}} \\
\midrule

\multirow{2}{*}{\textbf{Goma}}
&{OA(\%) } & 89.90 & 90.12 & 84.99 & 85.34 & 91.24 & 92.45 & \textcolor{blue}{\textbf{92.67}}  & \textcolor{red}{\textbf{94.21}} \\
&AA (\%) &91.30&91.66&87.57&84.70&92.59&93.10&\textcolor{blue}{\textbf{93.38}}&\textcolor{red}{\textbf{94.53}}\\

&{F1(\%) } & 29.92 & 30.55 & 24.19 & 24.88 & 31.97 & 31.86 & \textcolor{blue}{\textbf{32.42}}   & \textcolor{red}{\textbf{34.12}} \\
\midrule
\multirow{2}{*}{\textbf{Les Cayes}}
&{OA(\%) } & 83.73 & 81.54 & 86.12 & 76.91 & 85.14 & \textcolor{blue}{\textbf{91.88}} & 86.71   & \textcolor{red}{\textbf{93.03}} \\

&AA (\%) &85.91&85.04&88.49&78.02&86.17&\textcolor{red}{\textbf{91.65}}&87.55&\textcolor{blue}{\textbf{90.72}} \\

&{F1(\%) } & 27.87 & 25.66 & 24.34 & 21.29 & 29.22 & \textcolor{blue}{\textbf{31.74}} & 30.00  & \textcolor{red}{\textbf{35.10}} \\
\midrule
\multirow{2}{*}{\textbf{Hawaii}}
&{OA(\%) } & 70.19 & 70.58 & 66.08 & 68.78 & 72.23 & 70.25 & \textcolor{blue}{\textbf{72.38}}  & \textcolor{red}{\textbf{72.91}} \\

&AA (\%) &70.59&69.65&66.60&65.13&70.20&\textcolor{red}{\textbf{71.45}}&70.76 &\textcolor{blue}{\textbf{70.96}}\\

&{F1(\%) } & 22.73 & 24.67 & 18.98 & 20.09 & 27.60 & 24.15 & \textcolor{blue}{\textbf{27.90}}  & \textcolor{red}{\textbf{29.76}} \\

\midrule
\multirow{2}{*}{\textbf{La Palma}}
&{OA(\%) } & 87.45 & 87.66 & 79.94 & 80.67 & 88.05 & 82.70 & \textcolor{red}{\textbf{88.68}}  & \textcolor{blue}{\textbf{88.10}} \\

&AA (\%) &88.56&88.06&82.31&82.82&88.34&83.50&\textcolor{blue}{\textbf{88.63}}&\textcolor{red}{\textbf{88.74}} \\

&{F1(\%) } & 29.05 & 29.24 & 24.11 & 23.49 & 30.12 & 25.34 & \textcolor{blue}{\textbf{30.68}} & \textcolor{red}{\textbf{31.92}} \\
\midrule
\multirow{2}{*}{\textbf{Derna}}
&{OA(\%) } & \textcolor{blue}{\textbf{80.46}} & 79.94 & 76.75 & 77.37 & 80.35 & 80.14 & \textcolor{red}{\textbf{80.54}} & 80.24 \\

&AA (\%) &\textcolor{red}{\textbf{81.29}}&79.52&78.58&76.75&79.94&\textcolor{blue}{\textbf{81.14}}&80.91&79.89\\

&{F1(\%) } & 27.26 & 27.30 & 23.46 & 22.84 & 28.14 & 27.47 & \textcolor{blue}{\textbf{28.83}}   & \textcolor{red}{\textbf{32.63}} \\
\midrule
\multirow{2}{*}{\textbf{Marshall}}
&{OA(\%) } & 86.49 & 86.43 & 86.26 & 85.83 & 87.86 & \textcolor{blue}{\textbf{92.13}} & 88.24  & \textcolor{red}{\textbf{92.47}} \\

&AA (\%)  &88.19&88.64&87.94&86.59&89.32&\textcolor{blue}{\textbf{91.46}}&89.98&\textcolor{red}{\textbf{92.75}} \\

&{F1(\%) } & 28.84 & 29.07 & 27.55 & 24.94 & 30.48 & \textcolor{blue}{\textbf{31.86}} & 31.46 & \textcolor{red}{\textbf{33.99}} \\
\midrule
\textbf{Moulay}
&{OA(\%) } & 80.49 & 83.67 & 80.08 & 81.20 & 85.31 & \textcolor{red}{\textbf{91.67}} & 86.17   & \textcolor{blue}{\textbf{88.70}} \\

\textbf{Brahim} &AA (\%) &84.62&85.97&84.55&83.07&88.25&\textcolor{blue}{\textbf{92.80}}&89.16 &\textcolor{red}{\textbf{96.03}}\\

&{F1(\%) } & 24.65 & 25.32 & 23.51 & 22.84 & 27.38 & \textcolor{blue}{\textbf{29.69}} & 29.49  & \textcolor{red}{\textbf{31.90}} \\
\midrule
\multirow{2}{*}{\textbf{Antakya}}
&{OA(\%) } & 82.64 & 83.69 & 73.42 & 74.81 & 84.17 & \textcolor{red}{\textbf{88.40}} & 84.78  & \textcolor{blue}{\textbf{88.05}} \\

&AA (\%) &83.10&80.27&72.61&75.30&84.20&\textcolor{red}{\textbf{89.26}}&84.90&\textcolor{blue}{\textbf{89.12}} \\

&{F1(\%) } & 25.92 & 26.18 & 22.11 & 22.06 & 27.11 & \textcolor{blue}{\textbf{30.10}} & 28.25  & \textcolor{red}{\textbf{32.01}} \\

\midrule
\textbf{Average} &{OA(\%) } &83.30&83.59&80.49&80.02&84.76& \textcolor{blue}{\textbf{86.74}}&85.49 &\textcolor{red}{\textbf{87.68}}\\

\textbf{Across} &AA (\%) &84.21&83.77&82.30&80.42&85.10&\textcolor{blue}{\textbf{87.49}}&85.99&\textcolor{red}{\textbf{88.10}} \\

\textbf{Regions} & {F1(\%) } &27.15&27.38&23.87&22.94&28.88&29.15&\textcolor{blue}{\textbf{29.70}} &\textcolor{red}{\textbf{32.50}}\\

\bottomrule
\end{tabular}
    
\end{center}
\end{table*}

\subsection{Comparison Against the State-of-the-Art}
Extensive comparisons were conducted against a broad spectrum of recent methods: RSVQA \cite{lobry2020rsvqa}, RSIVQA \cite{zheng2021mutual}, CDVQA \cite{yuan2022change}, BiModal \cite{bazi2022bi}, SOBA \cite{wang2024earthvqa}, RSAdapter \cite{wang2024rsadapter}, and EarthVQANet \cite{wang2024earthvqanet}. However, with the exception of \cite{yuan2022change}, all the aforementioned methods have been proposed originally for RSVQA and not CDVQA, and hence only use a single scene as input, and not a bi-temporal pair. That is why, for the sake of fairness, their respective visual backbones have been provided as input with a concatenation of the pre- and post-event images. 

For all models, the training and validation splits were derived exclusively from the source domains as per the DG setting. 

\subsection{Quantitative Results}
As shown in Table \ref{tab:transposed_performance}, the proposed method, TCSSM, consistently outperforms a wide range of SOTA approaches across ten diverse geographic regions.


\textbf{Overall Performance:} On average, TCSSM achieves the highest OA, surpassing all competing models, including established CNN-based methods such as RSVQA and RSIVQA, Transformer-based methods such as SOBA, EarthVQANet, and RSAdapter. In terms of AA, TCSSM again leads, followed closely by RSAdapter. Similarly, in terms of F1, TCSSM attains the best performance. These findings suggest that the proposed models not only achieve high OA but also maintain robust and balanced detection capabilities across classes. It is also worth noting that the F1 score is relatively lower compared to OA and AA, primarily due to class imbalance, which causes the model to perform better on dominant classes while underrepresenting minority ones.



\textbf{Region-wise Analysis:} The superiority of the proposed methods is evident across multiple challenging locations. For instance, in Bata, TCSSM achieves the highest OA and F1-scores. In Beirut, the performance gap widens, as TCSSM surpasses the best baseline, RSAdapter, in F1-score.

In more challenging cases such as Hawaii, where baseline models perform relatively poorly, TCSSM demonstrates clear improvements, surpassing the next best model, EarthVQANet, across OA and F1 metrics. Similar trends are observed in regions such as La Palma, Marshall, and Moulay Brahim, where the proposed method consistently outperforms prior approaches across OA, AA, and F1-score.


\subsection{Qualitative Results}
\begin{figure*}[t]
\begin{center}
    
\begin{minipage}{0.23\textwidth}
    \includegraphics[width=\textwidth]{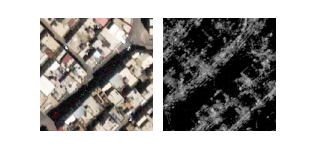}

    {\footnotesize
    \begin{description}[leftmargin=0pt, labelwidth=!,labelsep=0.3em, itemindent=0pt, font=\normalfont\bfseries]
        \item[Q:] Is there any visible damage in the post-disaster image?
        \item[GT:] No
        \item[RSVQA:] intact structures
        \item[RSIVQA:] spread throughout
        \item[CVDQA:] buildings that survived
        \item[BIModal:] partial damage
        \item[SOBA:] intact structures
        \item[RSAdapter:] No
        \item[EarthVQANet:] No
        \item[TCSSM (Ours):] No
    \end{description}}
\end{minipage}%
\hfill
\begin{minipage}{0.23\textwidth}
    \includegraphics[width=\textwidth]{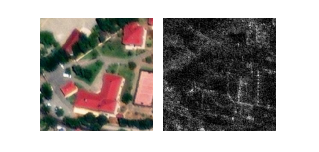}

    {\footnotesize
    \begin{description}[leftmargin=0pt, labelwidth=!,labelsep=0.3em, itemindent=0pt, font=\normalfont\bfseries]
        \item[Q:] what percentage of buildings
show partial damage?
        \item[GT:] 0\%
        \item[RSVQOA:] 0\%-10\%
        \item[RSIVQA:] No
        \item[CVDQA:] no damage
        \item[BIModal:] partial damage
        \item[SOBA:] 0\%-10\%
        \item[RSAdapter:] 0\%
        \item[EarthVQANet:] 0\%
        \item[TCSSM (Ours):] 0\%
    \end{description}}
\end{minipage}%
\hfill
\begin{minipage}{0.23\textwidth}
    \includegraphics[width=\textwidth]{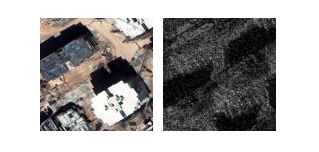}
    {\footnotesize
    \begin{description}[leftmargin=0pt, labelwidth=!,labelsep=0.3em, itemindent=0pt, font=\normalfont\bfseries]
        \item[Q:] is less than 10 \% of the area
damaged?
        \item[GT:] Yes
        \item[RSVQOA:] destroyed
        \item[RSIVQA:] equal
        \item[CVDQA:] minor damage
        \item[BIModal:] partial damage
        \item[SOBA:] Yes
        \item[RSAdapter:]Yes
        \item[EarthVQANet:] mostly intact
        \item[TCSSM (Ours):] Yes
    \end{description}}
\end{minipage}%
\hfill
\begin{minipage}{0.23\textwidth}
    \includegraphics[width=\textwidth]{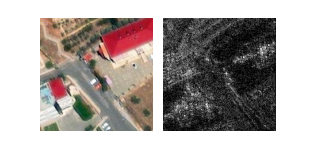}
    {\footnotesize
    \begin{description}[leftmargin=0pt, labelwidth=!,labelsep=0.3em, itemindent=0pt, font=\normalfont\bfseries]
        \item[Q:] is the area mostly intact or
mostly damaged?
        \item[GT:] mostly damaged
        \item[RSVQOA:] none
        \item[RSIVQA:] minor damage
        \item[CVDQA:] no damage
        \item[BIModal:] Yes
        \item[SOBA:] mostly damage
        \item[RSAdapter:] mostly damage
        \item[EarthVQANet:] 0\%
        \item[TCSSM (Ours):] mostly damaged
    \end{description}}
\end{minipage}%

\caption{Qualitative comparison of VQA model responses across various disaster-related questions.}
\label{fig:visu}
\end{center}

\end{figure*}

In terms of qualitative analysis, a range of disaster-related questions were posed, and the responses generated by the models under study were compared against the ground truth (Fig.~\ref{fig:visu}). It was observed that most of them provided inconsistent or incorrect answers, particularly in scenarios requiring fine-grained damage assessment. In contrast, responses generated by the proposed TCSSM showed a closer alignment with the annotations, especially for questions involving percentage estimation of partial damage and binary judgments about area integrity. 


\subsection{Ablation Study}

\subsubsection{\textbf{Question-based Performance Comparison}}

\begin{table*}[t]
\scriptsize
\begin{center}
\caption{Question-based performance comparison for the Bata region. The best result is highlighted in \textcolor{red}{red}, and the second-best in \textcolor{blue}{blue}.}
\label{tab:combined_question_based}
\resizebox{\textwidth}{!}{%
\begin{tabular}{llccccccc|c}
\toprule
\textbf{Location} & \textbf{Category} & \textbf{RSVQA} & \textbf{RSIVQA} & \textbf{CDVQA} & \textbf{BiModal} & \textbf{SOBA} &\textbf{RSAdapter} & \textbf{EarthVQANet} & \textbf{TCSSM} \\
&  & (TGRS'20)  & (TGRS'22)  & (TGRS'22) & (TGRS'22) & (AAAI'24)  & (TGRS'24)  & (ISPRS'24)  & (Ours) \\
\midrule
\multirow{10}{*}{\textbf{Bata}}
 & Comparative (\%)  &76.83 & 77.65& 78.07 & \textcolor{red}{\textbf{80.31}} & 78.34 &79.48& 78.88  & \textcolor{blue}{\textbf{80.06}} \\
 
 & Contextual (\%) &78.60 & 79.24& 80.83 & 80.83&80.24 &\textcolor{blue}{\textbf{80.97}}&\textcolor{red}{\textbf{81.43}}  & 66.19 \\
 
 & Damage Detection (\%)  &76.53 & 78.32 & 81.17 & 81.17&78.13& \textcolor{blue}{\textbf{83.01}}&79.64  &\textcolor{red}{\textbf{83.35}}  \\
 
 & Quantitative (\%) &63.55 & 64.84& 67.54 &68.03 & 65.76 &\textcolor{red}{\textbf{74.76}}&66.05 & \textcolor{blue}{\textbf{72.72}} \\
 
 & Recovery Assessment (\%) &80.26 & 80.37 &\textcolor{blue}{\textbf{99.78}} & 99.71& 86.36&\textcolor{blue}{\textbf{99.78}}&89.75  & \textcolor{red}{\textbf{99.83}}\\
 
 & Severity (\%) & 81.32 & 82.41 &\textcolor{red}{\textbf{97.69}} & \textcolor{red}{\textbf{97.69}}& 87.81&\textcolor{blue}{\textbf{97.65}} &91.27 & 97.57\\
 
 & Spatial (\%)& 82.83& 83.96& \textcolor{blue}{\textbf{88.42}} &88.10 &84.52 &\textcolor{red}{\textbf{88.62}}&86.74 &88.40 \\
 
 & Threshold (\%)& 84.10& 85.29 & 99.23 & 98.71&89.30&\textcolor{blue}{\textbf{99.35}} &93.28 & \textcolor{red}{\textbf{99.43}} \\

 & AA (\%) &78.00&79.01&86.59&\textcolor{blue}{\textbf{86.81}} &81.30&\textcolor{red}{\textbf{87.95}}&83.38&85.94 \\
 
 & OA (\%)  &82.63&82.76&84.53& 84.96&83.12&\textcolor{blue}{\textbf{86.47}}&83.28&\textcolor{red}{\textbf{86.87}} \\

\bottomrule
\end{tabular}%
}\end{center}
\end{table*}

Table \ref{tab:combined_question_based} presents the question category-based performances on the BrightVQA dataset's Bata region. The ablation study was conducted only for Bata, as similar performance trends were observed across all regions. 
More specifically, in the Recovery Assessment category, TCSSM attains the highest accuracy, slightly surpassing RSAdapter. Similarly, in the Damage Detection category, TCSSM outperforms the second-best RSAdapter. The AA and OA metrics further confirm the superior performance of the proposed model, TCSSM, achieving the highest scores among the compared methods. These results show the effectiveness and robustness of the proposed models in handling diverse question categories within the CDVQA task.

\subsubsection{\textbf{Language Bias}}

\begin{table}[h]
\scriptsize
\begin{center}
\caption{Effect of language bias on TCSSM for the Bata region. Best results (\%) are in bold.}
\label{tab:lanb}
\begin{tabular}{ccc|cccc}
\toprule
\textbf{Question} & \textbf{Vision} & \textbf{Description} & \textbf{Recall} & \textbf{Prec.} & \textbf{OA} & \textbf{F1} \\
\midrule
\cmark & \xmark & \xmark & 26.48 & 22.32 & 81.43 & 24.22 \\
\xmark & \cmark & \xmark & 10.12 & 7.65 & 36.05 & 8.71 \\
\xmark & \cmark & \cmark & 12.36 & 10.57 & 40.24 & 11.39\\
\cmark & \xmark & \cmark & 27.45 & 22.66 & 82.17 & 24.82 \\
\cmark & \cmark & \xmark & 26.53 & 22.49 & 81.68 & 24.34 \\
\midrule
\cmark & \cmark & \cmark & \textbf{29.76} & \textbf{29.58} & \textbf{86.87} & \textbf{29.66} \\
\bottomrule
\end{tabular}
\end{center}
\end{table}

Here, an ablation study was conducted by selectively removing key components - namely, the question input, vision inputs, and the geo-disaster-related description, thus assessing the relative importance and contribution of each input type, as well as the effects of their combinations on model performance. Results are being reported only for the Bata region, since similar patterns were observed across the remaining regions.


According to Table \ref{tab:lanb}, the configuration in which only the question input was retained resulted in a relatively high OA, with moderate recall and precision. This suggests that correct answers could be produced by relying primarily on question semantics, indicating the presence of language bias—that is, the ability to answer questions accurately without utilizing visual or descriptive inputs. This observation was also substantiated in \cite{deng2025words}, where it was shown that the majority of VQA models predominantly rely on textual information from the questions rather than visual content.

In contrast, when only vision or vision \& description modalities were used without the question, a sharp decline in performance was observed, particularly in recall and F1-score. These results confirm that, in the absence of the question input, accurate predictions could not be reliably generated.

Notably, when the question was combined with the description while excluding vision, a slight improvement over the question-only setting was observed, as reflected by an increase in F1-score. This suggests that textual descriptions can provide complementary context when visual features are unavailable.
Further improvements were observed when the question was combined with vision while excluding the description, yielding a higher F1-score. This indicates that visual features are more effectively leveraged when paired with the question than when paired solely with descriptions.


The best performance was achieved in the full model setting, where all three modalities were incorporated. In this configuration, both F1-score and OA reached their highest values, demonstrating that the comprehensive integration of question, vision, and description inputs contributes substantially to the model’s success.

\begin{table}[h]
\begin{center}
\caption{Effect of the number of TCSSM blocks on performance for the Bata region (using the validation split). The best results (\%) are in bold.}
\label{tab:num_layers_sel}
\begin{tabular}{lcccc|cc}
\toprule
\textbf{\# of} & \textbf{Recall} & \textbf{Precision} & \textbf{OA} & \textbf{F1} & \textbf{Params.} & \textbf{FLOPs} \\
\textbf{Blocks} & & & & &  (M) & (G)\\
\midrule
$L$ = 1&30.45  & 27.13 & 84.50 &28.69& \textbf{17.96} & \textbf{2.02} \\
$L$ = 2&  30.64 & \textbf{29.02} & \textbf{90.32} & \textbf{29.80} &18.66 &2.22\\  
$L$ = 3  & \textbf{31.34} & 27.99 & 89.02 & 29.57 &19.37 &2.41\\
 $L$ = 4 & 27.66 & 27.06 &83.38  & 27.35 &20.07&2.6 \\
\bottomrule
\end{tabular}
\end{center}
\end{table}

\subsubsection{\textbf{Number of Blocks}}
Table \ref{tab:num_layers_sel} presents the results of ablation experiments conducted on the Bata dataset (validation split) to evaluate the impact of varying the number of blocks in the model. 
It can be observed that performance varies, with the best results obtained using a 2-block configuration. This setting yields the highest precision, a strong F1-score, and a high OA, while also offering a favorable balance between predictive performance and computational complexity.


\begin{table*}[t]
\scriptsize
\begin{center}
\caption{Single DG results in terms of OA (\%), with Antakya as the source domain due to its large data volume. The best result per region is highlighted in \textcolor{red}{red}, and the second-best in \textcolor{blue}{blue}.}
\label{tab:SDG}
\resizebox{\textwidth}{!}{%
\begin{tabular}{lccccccc|cc}
\toprule
\textbf{Experiment}  & \textbf{RSVQA} & \textbf{RSIVQA} & \textbf{CDVQA} & \textbf{BiModal} & \textbf{SOBA} &\textbf{RSAdapter} & \textbf{EarthVQANet} & \textbf{TCSSM} \\

&   (TGRS'20)  & (TGRS'22)  & (TGRS'22) & (TGRS'22) & (AAAI'24)  & (TGRS'24)  & (ISPRS'24)  & (Ours) \\
\midrule

Antakya $\rightarrow$ Bata&84.72&85.10&83.84&84.96& 85.13&\textcolor{blue}{\textbf{85.95}}&85.32&\textcolor{red}{\textbf{87.59}}\\

Antakya $\rightarrow$ Beirut&89.58&88.41&89.78&89.74&88.35 &\textcolor{blue}{\textbf{90.60}}&89.70&\textcolor{red}{\textbf{90.94}} \\

Antakya $\rightarrow$ Goma&77.97&78.00&79.66&\textcolor{blue}{\textbf{85.68}}&83.78 &81.97&83.54&\textcolor{red}{\textbf{91.80}} \\

Antakya $\rightarrow$ Les Cayes&88.16&\textcolor{red}{\textbf{89.25}}&87.63&88.48&89.05 &88.65&\textcolor{blue}{\textbf{89.10}}& 88.50\\

Antakya $\rightarrow$ Hawaii&57.94&55.30&58.86&\textcolor{blue}{\textbf{63.31}}&61.38 &59.30&61.56&\textcolor{red}{\textbf{71.10}} \\

Antakya $\rightarrow$ La Palma&71.76&72.03&73.75&\textcolor{blue}{\textbf{79.90}}& 75.66&74.70&75.17& \textcolor{red}{\textbf{87.53}} \\

Antakya $\rightarrow$ Derna&74.42&75.13&74.22&75.58&76.21 &\textcolor{blue}{\textbf{76.47}}&76.21&\textcolor{red}{\textbf{81.74}} \\

Antakya $\rightarrow$ Marshall&79.65&76.17&80.98&\textcolor{blue}{\textbf{86.52}}&83.42 &83.31&83.76&\textcolor{red}{\textbf{91.07}} \\

Antakya $\rightarrow$ Moulay Brahim&78.01&77.93&77.87&80.08&79.93 &79.85&\textcolor{blue}{\textbf{80.12}}& \textcolor{red}{\textbf{85.90}}\\

\midrule
Avg. Across Regions  &78.02&77.48&78.51&\textcolor{blue}{\textbf{81.58}}& 80.32&80.08&80.50&\textcolor{red}{\textbf{86.24}} \\
\bottomrule
\end{tabular}%
}
\end{center}
\end{table*}

\subsubsection{\textbf{Single Domain Generalization}}

\begin{table*}[h]
\begin{center}

\caption{Effect on performance of different multi-temporal fusion strategies. the best result per region is highlighted in \textcolor{red}{red}, and the second-best in \textcolor{blue}{blue}.}
\label{tab:vqa_ablation}
\scriptsize

\begin{tabular}{lcccccccccc}
\toprule
\textbf{Location} &\textbf{Metric}& \textbf{Concat} & \textbf{Sum} & \textbf{Mul} & \textbf{Sub} & \textbf{Nsub} & \textbf{MCB} & \textbf{MUTAN} & \textbf{VisualBERT} & \textbf{MambaFusion}  \\

\midrule
\multirow{4}{*}{\textbf{Bata}}
&Recall(\%) &27.80&29.67&29.76&30.22&25.41&27.46&27.05&\textcolor{red}{\textbf{32.57}}&\textcolor{blue}{\textbf{31.54}}\\
&Precision(\%) &29.13&28.15&29.58&28.58&22.34&27.54&27.47&\textcolor{red}{\textbf{31.74}}&\textcolor{blue}{\textbf{29.82}}\\
&OA(\%)  &83.44&87.68&86.87&88.27&83.10&86.26&85.90 & \textcolor{red}{\textbf{89.68}}&\textcolor{blue}{\textbf{88.77}}\\
&F1 (\%) &28.44&28.89&29.66&29.37&23.77&27.47&27.28&\textcolor{red}{\textbf{32.18}}&\textcolor{blue}{\textbf{30.67}}\\

\midrule
\multirow{2}{*}{\textbf{Beirut}}
&OA(\%) &90.65&91.87&\textcolor{blue}{\textbf{92.31}}&91.67&82.54&90.37&89.83&91.41&\textcolor{red}{\textbf{92.50}} \\
&F1 (\%) &32.02&31.42&\textcolor{blue}{\textbf{34.00}}&30.90&23.45&31.56&30.56&31.65&\textcolor{red}{\textbf{34.13}}\\

\midrule
\multirow{2}{*}{\textbf{Goma}}
&OA(\%) &\textcolor{red}{\textbf{94.57}}&\textcolor{blue}{\textbf{94.25}}&94.21& 93.35&85.67&92.35&92.33&91.66&90.74\\
&F1 (\%)&33.71&\textcolor{blue}{\textbf{34.07}}&\textcolor{red}{\textbf{34.12}}&32.30&24.26&32.15&31.15&33.46&32.91 \\

\midrule
\multirow{2}{*}{\textbf{Les Cayes}}
&OA(\%) &93.08&\textcolor{red}{\textbf{94.14}}&93.03&\textcolor{blue}{\textbf{94.13}}&86.59&92.12&91.47 &92.86&90.98 \\
&F1 (\%) &32.64&33.34&\textcolor{blue}{\textbf{35.10}}&33.37&25.44&32.89&32.76&\textcolor{red}{\textbf{35.95}}&32.93 \\

\midrule
\multirow{2}{*}{\textbf{Hawaii}}
&OA(\%) &70.02&69.78&\textcolor{blue}{\textbf{72.91}}&68.34&66.07 &71.10&70.54&\textcolor{red}{\textbf{75.61}}&71.33\\
&F1 (\%) &23.48&22.80&\textcolor{blue}{\textbf{29.76}}&23.80&19.17&24.12&25.69&\textcolor{red}{\textbf{30.78}}&25.61 \\

\midrule
\multirow{2}{*}{\textbf{La Palma}}
&OA(\%)  &88.00&88.25&88.10&88.13&80.48&87.02&86.21&\textcolor{red}{\textbf{89.68}}&\textcolor{blue}{\textbf{89.56}}\\
&F1 (\%) &29.44&30.11&\textcolor{red}{\textbf{31.92}}&29.82&22.63&29.90&29.26&31.21&\textcolor{blue}{\textbf{31.23}}\\

\midrule
\multirow{2}{*}{\textbf{Derna}}
&OA(\%) &81.47&80.97&80.24&81.30&76.40&80.57&80.40&\textcolor{red}{\textbf{83.10}}&\textcolor{blue}{\textbf{82.06}} \\
&F1 (\%)&28.13&29.00&\textcolor{red}{\textbf{32.63}}&28.12&22.26&\textcolor{blue}{\textbf{30.43}}&29.37&30.20&29.15 \\

\midrule
\multirow{2}{*}{\textbf{Marshall}}
&OA(\%)  &90.70&90.73&\textcolor{red}{\textbf{92.47}}&\textcolor{blue}{\textbf{91.97}} &82.95&91.05&90.29&89.22& 88.75\\
&F1 (\%)  &31.76&\textcolor{blue}{\textbf{32.44}}&\textcolor{red}{\textbf{33.99}}&31.20 & 23.87&32.05&30.43&31.74&30.52\\

\midrule
\multirow{2}{*}{\textbf{Moulay Brahim}}
&OA(\%) &\textcolor{red}{\textbf{92.79}}&\textcolor{blue}{\textbf{90.78}}&88.70&88.62&82.98&89.12&88.97&90.25&89.78\\
&F1 (\%) &\textcolor{blue}{\textbf{31.44}}&30.39&\textcolor{red}{\textbf{31.90}}&29.82&23.76 &28.15&29.64&30.01&31.28\\

\midrule
\multirow{2}{*}{\textbf{Antakya}}
&OA(\%) &86.14&88.83&88.05&87.32&83.47&87.02&86.73&\textcolor{red}{\textbf{89.96}}&\textcolor{blue}{\textbf{88.95}}  \\
&F1 (\%) &28.74&\textcolor{blue}{\textbf{31.48}}&\textcolor{red}{\textbf{32.01}}&30.28&26.50&30.17&29.61&31.15&29.95\\

\midrule
\multirow{2}{*}{\textbf{Avg.Across Regions}}
&OA(\%) &87.08&\textcolor{blue}{\textbf{87.72}}&87.68&87.31&81.02&86.69& 86.26&\textcolor{red}{\textbf{88.34}}&87.34\\
&F1 (\%) &29.98&30.39&\textcolor{red}{\textbf{32.50}}&29.89&23.51&29.88&29.57&\textcolor{blue}{\textbf{31.83}}&30.83\\

\bottomrule
\end{tabular}
    
\end{center}
\end{table*}

To evaluate the robustness of the proposed method under different DG scenarios, a single-domain generalization setting was adopted. In this setting, Antakya was selected as the source domain due to its inclusion of the largest number of images among all available regions, thereby providing a strong foundation for model training. According to Table \ref{tab:SDG} across all target regions, TCSSM consistently outperforms all baselines. This result shows that the proposed model is capable of extracting domain-invariant features in the single-DG setting as well.


\subsubsection{\textbf{Effect on performance of multi-temporal fusion strategies}}

\begin{table}[h!]
\begin{center}
\caption{Comparison of inference time and Parameters for different fusion strategies.}
\label{tab:vqa_inference_time}
\begin{tabular}{c|cc}
\toprule
\textbf{Fusion Method} & \textbf{Inference Time (ms)} & \textbf{Params. (M)} \\
\midrule
Hadamard & 15 & 18.66 \\
VisualBERT~\cite{li2019visualbert} & 220 &36.89\\
MambaFusion~\cite{beedu2025mamba} & 244 & 34.27 \\
\bottomrule
\end{tabular}
\end{center}

\end{table}

To investigate the optimal strategy for fusing question and visual features before the answer prediction stage, several fusion operations were explored. Specifically, concatenation (Concat), element-wise summation (Sum), Hadamard product (Mul), subtraction (Sub), normalized subtraction (Nsub), multi-modal compact bilinear pooling (MCB)\cite{fukui2016multimodal}, multimodal Tucker fusion (MUTAN) \cite{ben2017mutan}, VisualBERT \cite{li2019visualbert}, and MambaFusion \cite{beedu2025mamba} were tested. 
According to Table \ref{tab:vqa_ablation}, the Hadamard product has consistently yielded strong performance across most metrics and regions, achieving the best average F1-score. The normalized subtraction approach underperformed across all regions. These results highlight the effectiveness of element-wise operations—especially multiplication—in capturing meaningful interactions between multi-modal features for VQA in disaster scenarios. 
Furthermore, to assess computational efficiency, inference times and parameters for Hadamard, VisualBERT, and MambaFusion were additionally compared (Table~\ref{tab:vqa_inference_time}). The Hadamard fusion was found to be the fastest, whereas VisualBERT was observed to be significantly slower, and MambaFusion was measured to be the slowest. These findings underscore the practicality of lightweight element-wise operations for time-sensitive disaster response applications.

\subsubsection{\textbf{Data Efficiency}}
\begin{figure}[t]
    \begin{center}
    \includegraphics[width=1.0\linewidth]{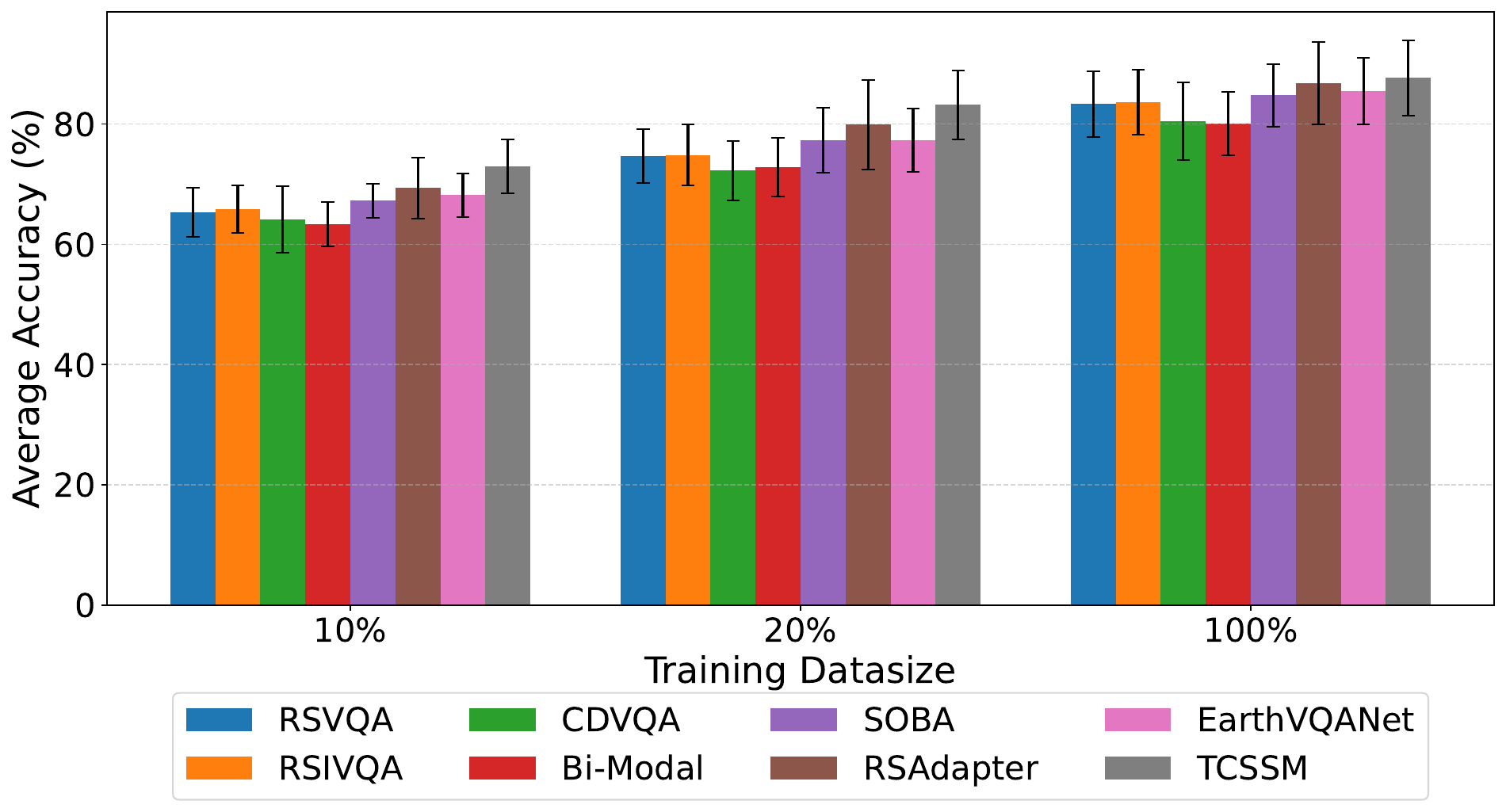}
    \caption{Average accuracy (\%) across all geographical regions of various VQA models evaluated with different training data sizes (10\%, 20\%, and 100\%).}
    \label{fig:ab_vis}
    \end{center}
\end{figure}

Finally, to evaluate the robustness of the proposed model under limited supervision, experiments were conducted using varying fractions of training data. This setting was motivated by the practical constraint that large-scale disaster datasets are not always readily available in real-world scenarios. Specifically, subsets containing 10\%, 20\%, and 100\% of the original training data were sampled and used for training. As illustrated in Fig.~\ref{fig:ab_vis}, 
TCSSM has shown strong resilience to data scarcity, maintaining competitive performance even when trained on only a small portion of the data. The observed stability across different training scales highlights the model’s capacity to generalize effectively under low-resource conditions, making it a suitable candidate for deployment in disaster response applications where labeled data is often limited or costly to obtain. Additionally, the inclusion of error bars in Fig.~\ref{fig:ab_vis} shows the consistency of TCSSM across multiple domains, indicating lower performance variance compared to competing methods. This further confirms the robustness and reliability of the proposed model when exposed to diverse disaster scenarios.


\subsection{Limitations}
Despite its strong performance, one limitation of the TCSSM framework is its potential sensitivity to language bias. Specifically, because the model relies heavily on English-language disaster descriptions, it may tend to focus disproportionately on the textual input, potentially neglecting or under-utilizing the rich visual information from the remote sensing images. This reliance on the question or textual cues can lead to decreased generalization when the language context changes or when visual features alone are crucial for accurate understanding.

\section{Conclusion} \label{sec:conc}


In conclusion, this study addresses domain shift in the CDVQA task by introducing the BrightVQA dataset and proposing the TCSSM framework, which
integrates bi-temporal visual data and geo-disaster textual descriptions to learn
domain-invariant representations through a novel state space model selection
mechanism, consistently outperforming existing methods across diverse domains
as shown in experimental results and ablation studies; future work will explore
additional domain shifts in question, answer distributions, and multi-sensor inputs to further enhance the framework's robustness and generalization to real-world scenarios.

\bibliographystyle{unsrt}  
\bibliography{ref}

\end{document}